\documentclass[letterpaper, 10pt, conference]{ieeeconf}
\IEEEoverridecommandlockouts
\usepackage{graphicx}
\usepackage{amsmath,amssymb}
\usepackage{booktabs}
\usepackage{multirow}
\usepackage{xcolor}
\usepackage[hidelinks]{hyperref}
\usepackage{cite}
\usepackage{tikz}
\usetikzlibrary{positioning,arrows.meta,fit,backgrounds}

\newcommand{\CoreUTwoMin}{1}
\newcommand{\CoreUTwoMax}{15}

\newcommand{\CoreGripMax}{30}
\newcommand{\CoreGsMax}{34}
\newcommand{\CoreWxMax}{34}
\newcommand{\CoreOvBase}{96}
\newcommand{\CoreOvArm}{37}
\newcommand{\CoreArmLossMax}{19}

\newcommand{\CoreYawMin}{18}
\newcommand{\CoreYawMax}{74}
\newcommand{\CoreRmMin}{87}
\newcommand{\CoreRmMax}{114}

\newcommand{\XDecLook}{0}
\newcommand{\XDecPhys}{43}
\newcommand{\XDecPhysLow}{46}

\newcommand{\PiDecLook}{2}
\newcommand{\PiDecPhys}{35}
\newcommand{\PiDecPhysLow}{36}

\newcommand{\GrDecLook}{72}
\newcommand{\GrDecPhys}{29}
\newcommand{\GrDecPhysLow}{14}

\newcommand{\TauDecLook}{49}
\newcommand{\TauDecPhys}{9}
\newcommand{\TauDecPhysLow}{4}

\newcommand{\GrForceP}{0.015}
\newcommand{\RmOrcMaxDiff}{3}
\newcommand{\OvRm}{87}
\newcommand{\OvOrc}{97}

\newcommand{\OvRmOrcP}{0.17}
\newcommand{\PiOrcNtenFifteen}{115}

\newcommand{\PiDocN}{391/400}
\newcommand{\PiCkptN}{365/400}
\newcommand{\PiDocPct}{97.75}
\newcommand{\PiCkptPct}{91.25}
\newcommand{\PiDocP}{7\times10^{-5}}

\newcommand{\TauWristRot}{114}
\newcommand{\PiWristRot}{63}
\newcommand{\XWristRot}{113}
\newcommand{\XCamRotTwo}{101}
\newcommand{\XCamRotTen}{10}
\newcommand{\XCamShiftTen}{56}

\newcommand{\PiWristRotFive}{83}
\newcommand{\PiWristRotTwenty}{23}
\newcommand{\XDiagRho}{-0.76}
\newcommand{\PiDiagRho}{-0.55}
\newcommand{\XDiagN}{63}
\newcommand{\TauDiagRho}{-0.61}
\newcommand{\TauDiagN}{49}

\newcommand{\PiDiagN}{54}
\newcommand{\GrDiagRho}{-0.54}
\newcommand{\GrDiagN}{49}
\newcommand{\OvDiagRho}{0.02}
\newcommand{\OvDiagN}{31}
\newcommand{\OvDiagDrop}{29}
\newcommand{\OvDiagCI}{[-0.38, 0.44]}
\newcommand{\ArmMiniMin}{14}
\newcommand{\ArmMiniMax}{19}
\newcommand{\ArmMiniBaseMin}{17}
\newcommand{\ArmMiniBaseMax}{19}
\newcommand{\ArmMiniOv}{8}
\newcommand{\ArmMiniOvBase}{17}
\newcommand{\AgRhoX}{-0.38}
\newcommand{\AgRhoMin}{-0.11}
\newcommand{\AgRhoMax}{-0.06}
\newcommand{\XEffNexec}{-24.9}

\newcommand{\XCfgMin}{46.7}
\newcommand{\XCfgMax}{97.5}
\newcommand{\XSeedSD}{0.71}
\newcommand{\PiEffNexec}{+5.4}

\newcommand{\OvLeRobot}{68}
\newcommand{\OvSeq}{78}
\newcommand{\OvMj}{88}
\newcommand{\XPmPhys}{37}
\newcommand{\PiPmPhys}{30}
\newcommand{\GrPmPhys}{15}
\newcommand{\TauPmPhys}{4}
\newcommand{\XCsLook}{0}
\newcommand{\XCsPhys}{44--54}
\newcommand{\PiCsLook}{0--2}
\newcommand{\PiCsPhys}{35--41}
\newcommand{\GrCsLook}{57--65}
\newcommand{\GrCsPhys}{28--33}
\newcommand{\TauCsLook}{44--48}
\newcommand{\TauCsPhys}{7--10}
\newcommand{\CsMaxP}{5.2\times10^{-5}}
\newcommand{\CtPandaN}{99}
\newcommand{\CtPandaRamekin}{6}
\newcommand{\CtFgN}{92}
\newcommand{\CtFgRamekin}{17}

\newcommand{\RM}{response matching}

\title{\LARGE \bf
Robot-to-Robot Transfer of Public VLA Checkpoints:\\ Policy-Specific Failure Modes and Training-Free Corrections
}

\author{Jae Hun Hwang$^{1,2,*}$, Yechan Seo$^{1,3,*}$, and Hong-In Won$^{1,4}$%
\thanks{$^{*}$Equal contribution.}%
\thanks{$^{1}$Manufacturing AI Research Center, Korea Institute of Industrial Technology (KITECH), Incheon, Republic of Korea.}%
\thanks{$^{2}$Department of Artificial Intelligence Convergence, Hanyang University, Seoul, Republic of Korea.}%
\thanks{$^{3}$Department of Intelligent Robotics, Sungkyunkwan University, Suwon, Republic of Korea.}%
\thanks{$^{4}$HYU-KITECH Joint Department, Graduate School, Hanyang University, Seoul, Republic of Korea.}%
\thanks{Corresponding author: Hong-In Won (luvhayym@kitech.re.kr).}}

\begin{document}

\maketitle
\thispagestyle{empty}
\pagestyle{empty}

\begin{abstract}
Changing a robot's gripper can reduce manipulation success even when the task stays the same. We tested public checkpoints from five vision-language-action (VLA) policies in LIBERO simulation. Swapping the gripper's appearance and its physics separately revealed which change caused the larger loss for each policy. Across 12 original policy--gripper pairs, that comparison identified the better of two corrections: changing the wrist image or adjusting the gripper setup. One of four labels reversed on replication, so direct trials remain important when choosing a correction. For a rotated base frame, six probe motions let us calibrate commands for four relative-action policies without retraining. At $15^\circ$ yaw, successes rose from \CoreYawMin--\CoreYawMax{} to \CoreRmMin--\CoreRmMax{} per 120 episodes. Changes in wrist-image features ranked visual risk before execution, but missed failures caused by gripper physics. Two undocumented OpenVLA evaluation settings together changed success by 20 percentage points.
\end{abstract}


\section{Introduction}
A laboratory that downloads a released vision-language-action (VLA) checkpoint and runs the authors' inference code on a task the policy was trained for can still obtain near-zero success. Two families of explanations compete: the \emph{evaluation pipeline} (inference arguments, preprocessing, seeds)~\cite{choi2026vlaeval,wang2026industrialvlabench} and the \emph{deployment setup} (gripper, arm, camera, base frame, controller). Robustness benchmarks perturb cameras, lighting and initial states~\cite{fei2025liberoplus,zhou2025liberopro}, and cross-embodiment methods repaint or mask the robot, usually for policies the authors trained themselves~\cite{chen2024mirage,piseno2026cloak,yan2026zeta}. Here we compare released checkpoints under controlled hardware swaps and evaluate corrections that leave policy weights fixed.

In \emph{robot-to-robot transfer}, a policy performs the same task after a change in the arm, gripper, camera placement, base frame or controller~\cite{oxe2023openx}. We held the scene fixed to isolate hardware effects and varied lighting in a separate X-VLA check. Existing benchmarks examine changes in background, texture and lighting~\cite{fei2025liberoplus,zhou2025liberopro}.

\begin{figure*}[t]
\centering
\includegraphics[width=\textwidth]{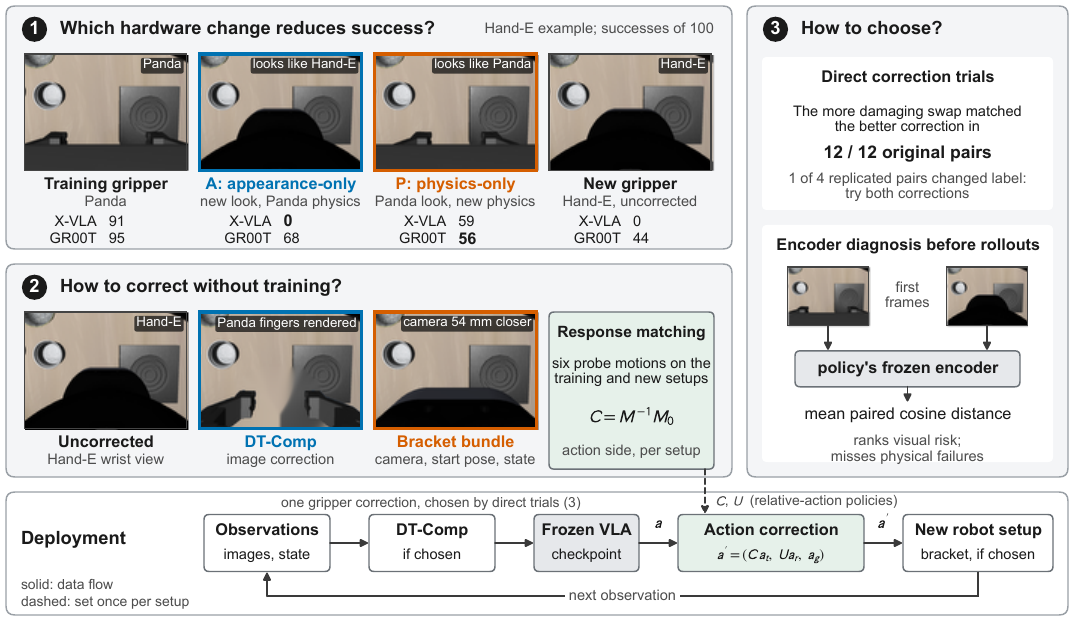}
\caption{Overview, organised by the three questions of the study. (1) Wrist-camera views of the training gripper, the appearance-only cell A, the physics-only cell P and the new gripper, with successes of 100 on LIBERO-Spatial for Hand-E; bold marks the lower of A and P for each policy. (2) Training-free corrections: DT-Comp repaints the wrist image using gripper masks and depth from the simulator, and the bracket bundle translates the wrist camera toward the fingers and matches the start pose and gripper-state range (Sec.~\ref{sec:corrm}); response matching calibrates the action correction per setup (Sec.~\ref{sec:rm}). (3) Candidate gripper corrections are compared by direct trials; the encoder shift between first frames of the training and new setups ranks visual risk before rollouts (Sec.~\ref{sec:diagm}). Bottom: at deployment DT-Comp acts on the observations, the bracket is part of the setup, and $U$ is the orthogonal polar factor of $C$. The modules are evaluated separately; diagnosis stays outside deployment.}
\label{fig:overview}
\end{figure*}

Which hardware changes reduce success? Can a laboratory recover performance with the policy weights fixed, and how should it choose a correction? We address these questions in LIBERO~\cite{liu2023libero}, starting from each policy's reproduced evaluation (Table~\ref{tab:repro}). For grippers, we vary appearance and physics separately. We also vary the arm, cameras, base frame and controller gain (Fig.~\ref{fig:overview}).

We compare two gripper corrections: compositing the training gripper into the wrist image and changing the camera bracket together with the gripper settings. For errors in commanded motion, we calibrate the new robot's response against the training robot's response.

Our contributions are:
\begin{enumerate}
\item We identify evaluation-pipeline effects and policy-specific hardware loss patterns through checkpoint reproduction and controlled swaps (Sec.~\ref{sec:eval}, Sec.~\ref{sec:grippers}, Sec.~\ref{sec:corr}).
\item We propose training-free gripper corrections using DT-Comp image compositing and a bundled bracket intervention, and compare their performance across policies and grippers (Sec.~\ref{sec:corr}).
\item We propose \RM{}, a closed-form action correction, and find that short execution absorbs tested gain errors but not frame errors (Sec.~\ref{sec:action}).
\item We propose rollout-free encoder diagnosis and find that wrist-view shift ranks visual risk but misses physical failures (Sec.~\ref{sec:diag}).
\end{enumerate}

\section{Related Work}
\textbf{VLA policies and benchmarks.} We study OpenVLA~\cite{kim2024openvla}, $\pi_{0.5}$~\cite{intelligence2025pi05}, X-VLA~\cite{zheng2025xvla}, GR00T~\cite{nvidia2025gr00tn1} and $\tau_0$-VLA~\cite{cai2026tau0vla} on LIBERO~\cite{liu2023libero}, using LeRobot~\cite{cadene2026lerobot} and the authors' released evaluation code.

\textbf{Robustness evaluation.} LIBERO-Plus and LIBERO-PRO perturb cameras, lighting, language and initial states~\cite{fei2025liberoplus,zhou2025liberopro}; multi-axis studies combine perturbations~\cite{sawada2026beyondsingleaxis}, and evaluation-pipeline audits show that settings change reported numbers~\cite{choi2026vlaeval}. We add embodiment-level factors built from official hardware models and compare success losses under controlled swaps.

\textbf{Cross-embodiment visual transfer.} Mirage cross-paints the robot without training~\cite{chen2024mirage}; Cloak retrains with the end-effector masked~\cite{piseno2026cloak}, consistent with the appearance-led losses we observe for two policies; ZETA reports little sensitivity to logos and colour for diverse-embodiment pretraining~\cite{yan2026zeta}. In the original correction trials, image correction outperforms the bracket for appearance-led policy--gripper pairs; this association does not hold for every replicated label.

\textbf{Action alignment.} Self-adaptive VLAs and TeAR adapt actions with post-training~\cite{zhang2026selfadaptivevla,sagar2026tear}. Response matching instead calibrates the command-to-motion response without training; we compare it with uncorrected execution, short execution and an exact-yaw oracle.

\textbf{Failure prediction.} Runtime detectors~\cite{omahoney2026distal,seligmann2026vlafail,romer2025fiper} and label-trained success predictors~\cite{huang2026vlaconf} operate per episode. We rank setups before deployment without rollouts on the new setup.

\section{Setup}\label{sec:setup}
\textbf{Policies.} X-VLA (\texttt{xvla-libero}) outputs absolute end-effector (EEF) targets in 30-step chunks. $\pi_{0.5}$ (\texttt{pi05\_libero\_finetuned}), GR00T N1.7 (\texttt{gr00t17-lerobot-libero\_spatial-640}) and $\tau_0$-VLA (\texttt{tau-0-vla-libero}) output relative EEF deltas from wrist and third-person images and proprioception, in chunks of 50 (stored default; 10 documented), 16, and 10, respectively; $\tau_0$ executes 8 actions per chunk. OpenVLA, a widely used open-weight baseline and the starting point of later work~\cite{kim2025openvlaoft}, represents the earlier design: one checkpoint per suite outputs single-step deltas from the third-person image only, without proprioception, evaluated in bfloat16 precision. For $\tau_0$, the authors' model server runs unchanged and our client builds requests exactly as their LIBERO client.

\textbf{Policy-specific evaluation environments.} Seeding and simulator version change the numbers (Sec.~\ref{sec:eval}), so each policy runs as its authors evaluated it: LeRobot seeding with MuJoCo~\cite{todorov2012mujoco} 3.8.1 for X-VLA, $\pi_{0.5}$ and GR00T; the original LIBERO scene sequence with MuJoCo 3.1.6 for OpenVLA Spatial/Object and 3.8.1 for Goal/Long; the original LIBERO with a fixed per-episode seed and MuJoCo 3.2.3 for $\tau_0$, all with robosuite~\cite{zhu2020robosuite} 1.4.0. Each comparison uses the same policy on both setups. For OpenVLA and $\tau_0$, first frames from our evaluation runner are pixel-identical to those of the authors' code, and for $\tau_0$ the two agree on 196 of 200 episodes. Table~\ref{tab:repro} shows that each policy's reproduced average is at least its reported average over the compared suites; the GR00T comparison covers the Spatial suite.

\begin{table}[t]
\centering
\caption{Reproduction of published LIBERO success (\%, Spatial/Object/Goal/Long, or average) on each policy's original evaluation environment. $\tau_0$: 500 episodes per suite; $\pi_{0.5}$: 100; OpenVLA: 50; X-VLA: 5 seeds $\times$ 400; GR00T: Spatial, 500.}
\label{tab:repro}
\scriptsize
\setlength{\tabcolsep}{3pt}
\resizebox{\linewidth}{!}{\begin{tabular}{@{}l l r r@{}}
\toprule
Policy & Evaluation stack & Ours (\%) & Reported (\%) \\
\midrule
$\pi_{0.5}$ & LeRobot seeding, MuJoCo 3.8.1 & 97/99/99/96 & 97/99/98/96 \\
X-VLA & LeRobot seeding, MuJoCo 3.8.1 & 97.2 & 93 \\
OpenVLA & sequential seeding, MuJoCo 3.1.6$^\dagger$ & 88/88/80/54 & 84.7/88.4/79.2/53.7 \\
GR00T N1.7 & LeRobot seeding, MuJoCo 3.8.1 & 95.2 & 95 \\
$\tau_0$-VLA & fixed per-episode seed, MuJoCo 3.2.3 & 98/99/98/96 & 97.4/98.8/98.2/95 \\
\bottomrule
\end{tabular}
}
\\[2pt]
\parbox{\linewidth}{\scriptsize $^\dagger$Goal and Long with MuJoCo 3.8.1. Reported values: LeRobot documentation~\cite{lerobotdocs2026} for $\pi_{0.5}$, X-VLA and GR00T; the OpenVLA paper~\cite{kim2024openvla}; the $\tau_0$-VLA release~\cite{sii2026tau0libero}.}
\end{table}

\textbf{Setups.} We used official gripper models: OnRobot 2FG7 and RG2 and SCHUNK EGK40 from manufacturer Unified Robot Description Format (URDF) files and meshes; Robotiq Hand-E and DH PGC-50-35 from manufacturer computer-aided design (CAD) models for the visuals and datasheet dimensions for the physics; ALOHA and WidowX from MuJoCo Menagerie~\cite{zakka2022menagerie}. Industrial grippers share the Panda's contact parameters (ALOHA and WidowX keep Menagerie's); the 2FG7's collision pad follows its datasheet, and its housing and fingertip collision geometry uses bounding boxes of the official meshes. The policies command opening and closing; the simulator sets the force. We used rated forces for industrial grippers and estimated the research grippers' forces from their drive specifications (see Gripper force settings in the supplement).

\begin{figure}[t]
\centering
\includegraphics[width=\linewidth]{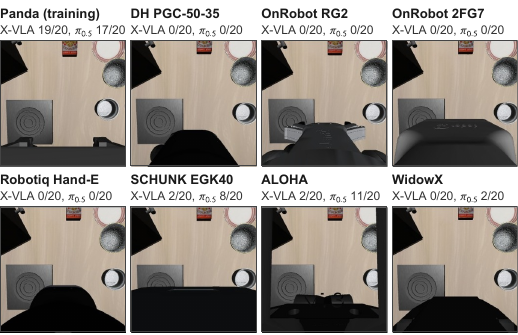}
\caption{First wrist-camera frame with the training gripper (Panda) and other grippers mounted, with successes from the diagnosis sweep (LIBERO-Spatial, 20 episodes). The Panda fingers occupy a small area of the image; industrial housings occupy the lower image. The Panda frame uses a different initial state.}
\label{fig:wrist}
\end{figure}

Arms (UR5e, RB5-850e, Doosan M1013) come from manufacturer descriptions, with the start pose of the tool centre point (TCP) matched by inverse kinematics before LIBERO's 10 settling steps, after which the TCP differs from the Panda's by at most 14\,mm and $6^\circ$ (median 6\,mm); the arm-swap results include this residual. Gripper swaps keep LIBERO's initial joint values, so that only the gripper differs from training; the start TCP moves by the length difference (2FG7: +42\,mm). We also vary camera poses, the base-frame yaw, the gain of the operational-space controller (OSC) and lighting. \emph{Appearance-only swaps} hide one gripper's visual geometry and attach the other's official meshes at the flange, so the wrist camera sees exactly the mounted product while the physics stays unchanged (Fig.~\ref{fig:wrist}).

\textbf{Protocol.} Core setups, tolerance and calibration tests use 4 suites $\times$ 10 tasks $\times$ 3 initial states (120 episodes); GR00T and OpenVLA use their per-suite checkpoints. The broader setup sweep for the diagnosis uses LIBERO-Spatial, 10 tasks $\times$ 2 initial states (20 episodes). We report Fisher exact tests, Wilson intervals and Spearman correlations with bootstrap intervals. Per-episode results, cell configurations and analysis scripts of the correction experiment are released with the paper; the remaining code and results will be released.

\section{Methods}
\subsection{Appearance/physics decomposition and training-free gripper corrections}\label{sec:corrm}
The appearance-only swap of Sec.~\ref{sec:setup} defines the decomposition cells: A uses target-gripper appearance on Panda physics, whereas P uses Panda appearance on target-gripper physics (Fig.~\ref{fig:overview}, panel~1). A policy--gripper pair is appearance-led if A has fewer successes than P and physics-led if P has fewer than A; a tie selects neither. This decomposition label indicates DT-Comp for appearance-led pairs and the bracket for physics-led pairs. Decomposition requires rollouts on initial states disjoint from correction trials; implementation settings are given in the supplementary section ``Image-correction settings'', and the split is described in the supplementary material.
DT-Comp (digital-twin compositing) changes the wrist image supplied to the frozen policy (Fig.~\ref{fig:overview}, panel~2). The target-gripper mask comes from simulator depth differences with and without target-gripper visuals and is filled by Telea's image-inpainting algorithm. A twin matches the arm joints and maps measured pad distance to the Panda's range, renders only Panda finger links, and pastes them where their depth is less than the gripper-free scene depth. DT-Comp uses simulator-privileged masks, camera parameters and gripper-free scene depth. GripSwap inpaints an offline target mask with the same inpainting algorithm and pastes an offline Panda template selected by mapped gripper opening.

The bracket bundles initial tool centre point alignment, gripper-state normalisation to the Panda range and wrist-camera translation toward the fingers: +42.09\,mm for 2FG7, +53.6\,mm for Hand-E and +23.8\,mm for PGC-50-35. The bundle retains target appearance, contact geometry and rated-force physics. The combination applies DT-Comp after the bracket, translating the twin camera by the same offset.

\subsection{Response matching}\label{sec:rm}
A relative-action command $a=(a_t, a_r, a_g)$ specifies incremental end-effector translation and rotation in the training robot's frame and scale, together with a gripper command. A robot with a rotated base frame or a different controller gain realises the same command differently, and the error accumulates over an open-loop chunk. We model the command-to-motion map of the translation channel as linear over the probe horizon.

\textbf{Probing.} From a fixed start state we send six probes $c_i \in \{\pm\delta_1 e_1, \pm\delta_2 e_2, \pm\delta_3 e_3\}$, each held for $k$ control steps, where $e_j$ is the $j$-th coordinate unit vector and $\delta_j$ is the corresponding translation-action standard deviation from the checkpoint's normaliser. Each probe starts from the same reset state. With $\Delta p_i$ the realised displacement of the robot's reported end-effector position after probe $i$, the map is the least-squares fit
\begin{equation}
M = \arg\min_{M'} \textstyle\sum_{i=1}^{6} \lVert \Delta p_i - M' c_i \rVert^2 .
\end{equation}
$M_0$ is estimated with the same probes once on the training embodiment.

\textbf{Correction.} A command $a_t$ produces $M_0 a_t$ on the training embodiment and $M a_t$ on the new one, so sending $C a_t$ with
\begin{equation}
C = M^{-1} M_0
\end{equation}
matches the training response within the fitted linear model, provided $M$ is invertible. Rotation deltas are corrected with the orthogonal factor $U$ of the polar decomposition $C = U P$, which carries a frame rotation but not a gain change; the gripper command is unchanged. The probe duration $k \in \{2,5,10\}$ was selected once on a tuning split of $\pi_{0.5}$ ($k{=}10$) and held fixed for the other policies; $\delta_j$ is fixed by the dataset. Calibration takes six motions of $k$ steps per setup.

\subsection{Rollout-free diagnosis}\label{sec:diagm}
A few still images of the new setup pass through the policy's own preprocessing and vision encoder (PaliGemma-SigLIP for $\pi_{0.5}$, Florence-2 for X-VLA, the fused DINOv2--SigLIP projector for OpenVLA, Qwen3-VL for GR00T, Qwen3.5-VL for $\tau_0$). Tokens are mean-pooled per camera $v$ into $f_v(x)$. For images $x^s_j$ and $x^0_j$ from setup $s$ and the training setup at matched task and initial state $j$, the shift is the mean paired cosine distance,
\begin{equation}
d_v(s) = \frac{1}{J}\sum_{j=1}^{J} \Bigl(1 - \cos\bigl(f_v(x^{s}_{j}), f_v(x^{0}_{j})\bigr)\Bigr),
\label{eq:shift}
\end{equation}
computed on the first frame of $J{=}20$ episodes without rollouts or labels. In leave-one-setup-out evaluation, a threshold fitted to the other setups' outcomes predicts a significant drop, and the camera with the larger standardised shift is identified as the affected view. The probe of Sec.~\ref{sec:rm} adds action-side signals: the rotation angle of the polar factor of $C$ and the log singular values, flagged when above the training-vs-training noise floor.


\section{Results}
\subsection{Unreported evaluation settings move the same checkpoint}\label{sec:eval}

\begin{figure*}[t]
\centering
\includegraphics[width=\textwidth]{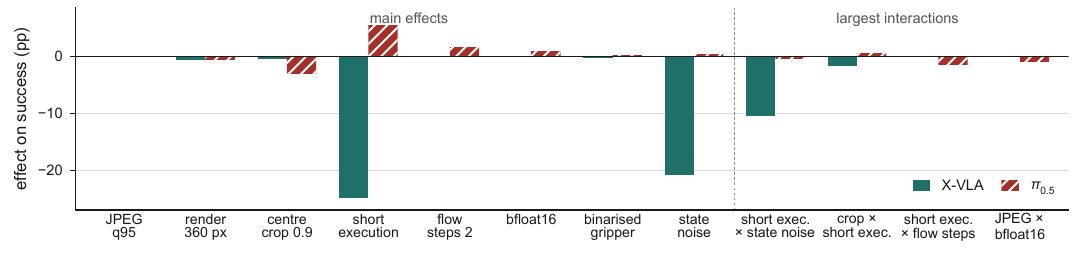}
\caption{Main effects (pp) and interactions for X-VLA and $\pi_{0.5}$ in a $2^{8-2}$ design with 120 episodes per configuration. Each factor switches from its reference level to JPEG quality 95, 360\,px render, centre crop 0.9, short execution (30$\to$5 steps for X-VLA, 50$\to$5 for $\pi_{0.5}$, from the chunk length stored in each checkpoint), 2 flow steps, bfloat16, binarised gripper command or state noise 0.3 standard deviations; the four interactions shown have the largest absolute effects over the two policies.}
\label{fig:factorial}
\end{figure*}

With the documented command our evaluation runner reproduces the official $\pi_{0.5}$ evaluation (\PiDocN, \PiDocPct\%); with the 50-step chunks stored in the checkpoint's own configuration, the same checkpoint scores \PiCkptN{} (\PiCkptPct\%, Fisher $p=\PiDocP$). For OpenVLA on LIBERO-Spatial, LIBERO places fixtures with a fresh random draw at every reset that the initial-state file does not pin. Re-seeding every episode (LeRobot-style) gives \OvLeRobot\%, the authors' sequential seeding \OvSeq\%, and additionally matching their simulator version (MuJoCo 3.1.6, which renders some textures differently from 3.8.1 with identical assets) \OvMj\%, at the published level (84.7\%) or above. Neither setting is stated in the corresponding papers. A $2^{8-2}$ factorial over eight pipeline settings (64 configurations $\times$ 120 episodes per policy; Fig.~\ref{fig:factorial}) shows that execution length dominates with opposite signs: shortening it changes X-VLA by $\XEffNexec$ percentage points (pp) and $\pi_{0.5}$ by $\PiEffNexec$\,pp. Configurations span \XCfgMin--\XCfgMax\% for X-VLA against a seed standard deviation of \XSeedSD\,pp. 

\subsection{Gripper swaps reveal policy-specific loss patterns}\label{sec:grippers}

Table~\ref{tab:core} summarises the five policies on the core setups over all four suites. Every industrial gripper drops every policy far below its nominal level (at most \CoreGripMax{} of 120 successes for any policy and gripper), and the UR5e + 2FG7 combination gives \CoreUTwoMin--\CoreUTwoMax{} of 120 across the five. Each gripper lifts the bowl in a scripted grasp, so the losses do not come from an inability to perform this grasp.

\begin{table}[t]
\centering
\caption{Success (of 120) on core setups, 4 suites $\times$ 10 tasks $\times$ 3 initial states. Each policy in its original evaluation environment (GR00T and OpenVLA with their per-suite checkpoints). n/a: not defined for the policy (OpenVLA has no wrist camera; X-VLA outputs absolute targets, while response matching corrects relative actions).}
\label{tab:core}
\scriptsize
\setlength{\tabcolsep}{3.5pt}
\resizebox{\linewidth}{!}{\begin{tabular}{@{}lrrrrr@{}}
\toprule
Setup & X-VLA & $\pi_{0.5}$ & OpenVLA & GR00T N1.7 & $\tau_0$-VLA \\
\midrule
Training setup (Panda gripper) & 116 & 108 & 96 & 113 & 116 \\
\addlinespace
OnRobot 2FG7 & 3 & 11 & 3 & 14 & 15 \\
DH PGC-50-35 & 5 & 8 & 1 & 30 & 12 \\
Robotiq Hand-E & 0 & 3 & 1 & 20 & 4 \\
\addlinespace
Panda physics, 2FG7 look & 6 & 8 & 2 & 35 & 40 \\
2FG7 physics, Panda look & 28 & 48 & 5 & 27 & 15 \\
\addlinespace
UR5e arm (Panda gripper) & 106 & 88 & 37 & 99 & 98 \\
UR5e + 2FG7 & 1 & 9 & 3 & 15 & 5 \\
2FG7, wrist camera +6\,cm & 34 & 21 & n/a & 12 & 19 \\
2FG7, GripSwap repainting & 34 & 28 & n/a & 34 & 27 \\
\addlinespace
Base frame yaw $15^\circ$ & 114 & 25 & 18 & 55 & 74 \\
\quad + response matching & n/a & 110 & 87 & 113 & 114 \\
\quad + oracle frame inversion & 116 & 108 & 97 & 110 & 116 \\
Controller gain $\times0.5$ & 117 & 23 & 82 & 104 & 115 \\
\bottomrule
\end{tabular}
}
\end{table}

\begin{figure*}[t]
\centering
\includegraphics[width=\textwidth]{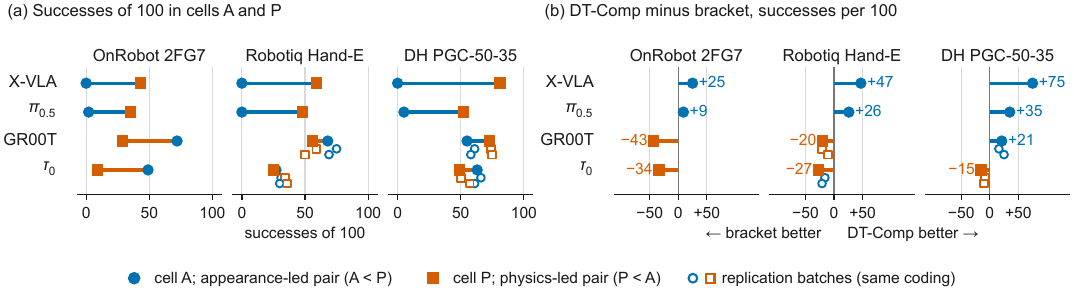}
\caption{(a) Successes of 100 on LIBERO-Spatial in the appearance-only cell A (blue circles) and the physics-only cell P (orange squares); the connecting line takes the colour of the lower cell. (b) DT-Comp minus bracket successes per 100, coloured by the label of the pair. Filled markers show original runs; open markers show replication batches with their own labels. GR00T changes from physics-led on Hand-E to appearance-led on PGC-50-35; for $\tau_0$ on Hand-E, both replication batches reverse the original label while the bracket stays better. Fig.~\ref{fig:overview} shows the wrist views of the cells and corrections.}
\label{fig:grip}
\end{figure*}

The appearance/physics swaps separate two groups (Table~\ref{tab:decomp}, 100 episodes per cell). With Panda physics rendered as a 2FG7, X-VLA and $\pi_{0.5}$ fail almost completely (\XDecLook{} and \PiDecLook{} of 100), and with 2FG7 physics rendered as a Panda they partly recover (\XDecPhys{} and \PiDecPhys): the appearance-only swap costs more successes. A gripper that occludes \emph{less} of the wrist image than the Panda (PGC-50-35) still fails, and painting the industrial grippers in the Panda's colours does not help, so neither reduced occlusion nor the tested colour change resolves it. GR00T and $\tau_0$ show the opposite pattern: they retain \GrDecLook{} and \TauDecLook{} successes with 2FG7 appearance on Panda physics but only \GrDecPhys{} and \TauDecPhys{} with Panda appearance on 2FG7 physics. OpenVLA, which sees no wrist view, fails both ways (Table~\ref{tab:core}). The three-gripper decomposition cells of the correction experiment (Sec.~\ref{sec:corr}) show the same appearance-led grouping for X-VLA and $\pi_{0.5}$ and a gripper-dependent label for GR00T (Fig.~\ref{fig:grip}(a)). Gripper length enters both cells: the wrist camera sits on the flange, so with 2FG7 physics it is 42\,mm farther from the object at grasp, and with 2FG7 appearance the rendered fingertips lie 42\,mm beyond the pads that touch it. P measures the combined effect of geometry, length, contact and camera distance at grasp. Varying that distance independently of gripper length requires camera remounting. For GR00T, non-pad geometry contacts the scene in \CtPandaN/100 Panda and \CtFgN/100 2FG7-physics episodes, mostly at the target bowl, with the 2FG7 bounding-box housing contacting the neighbouring ramekin in \CtFgRamekin{} versus \CtPandaRamekin{} episodes. X-VLA zeroes its gripper proprioception, so its physical deficit does not come from the gripper-width input. Matching the start TCP to the Panda's by inverse kinematics and mapping the gripper state onto the Panda's range gives no significant recovery (GR00T \GrDecPhys$\to$\GrPmPhys, $\tau_0$ \TauDecPhys$\to$\TauPmPhys, X-VLA \XDecPhys$\to$\XPmPhys, $\pi_{0.5}$ \PiDecPhys$\to$\PiPmPhys{} of 100; pre-registered test). The ordering of the two cells is also robust to the contact model: with friction or the contact time constant scaled by $\times0.9$ or $\times1.1$ on every geom, robosuite's default randomisation range, the appearance-only/physics-only cells give \XCsLook/\XCsPhys{} for X-VLA, \PiCsLook/\PiCsPhys{} for $\pi_{0.5}$, \GrCsLook/\GrCsPhys{} for GR00T and \TauCsLook/\TauCsPhys{} for $\tau_0$ (pre-registered, 16 one-sided Fisher tests, all significant after Holm, $p \le \CsMaxP$).

\begin{table}[t]
\centering
\caption{Appearance/physics decomposition with the OnRobot 2FG7, LIBERO-Spatial, 100 episodes per cell. Top: grippers at their rated grip force. Bottom: the 2FG7 closing at about 20\,N, the lower end of its configurable range.}
\label{tab:decomp}
\scriptsize
\setlength{\tabcolsep}{4pt}
\resizebox{\linewidth}{!}{\begin{tabular}{@{}lrrrr@{}}
\toprule
Setup & X-VLA & $\pi_{0.5}$ & GR00T N1.7 & $\tau_0$-VLA \\
\midrule
Training setup (Panda gripper) & 89 & 82 & 94 & 97 \\
OnRobot 2FG7 & 0 & 1 & 16 & 13 \\
Panda physics, 2FG7 look & 0 & 2 & 72 & 49 \\
2FG7 physics, Panda look & 43 & 35 & 29 & 9 \\
UR5e + 2FG7 & 0 & 0 & 10 & 10 \\
\midrule
\multicolumn{5}{@{}l}{\emph{2FG7 closing at about 20\,N (shared Panda finger gain)}} \\
OnRobot 2FG7 & 0 & 1 & 9 & 13 \\
2FG7 physics, Panda look & 46 & 36 & 14 & 4 \\
\bottomrule
\end{tabular}
}
\end{table}

\textbf{Grip force.} Because the policies command opening and closing, the grip force is a deployment setting. Lowering the 2FG7 from its rated 140\,N to about 20\,N, the bottom of its configurable range, leaves X-VLA and $\pi_{0.5}$ nearly unchanged (\XDecPhys{}$\to$\XDecPhysLow, \PiDecPhys{}$\to$\PiDecPhysLow), leaves $\tau_0$ near its floor (\TauDecPhys{}$\to$\TauDecPhysLow, not significant), but lowers GR00T's 2FG7-physics/Panda-appearance cell from \GrDecPhys{} to \GrDecPhysLow{} (Fisher $p=\GrForceP$; bottom of Table~\ref{tab:decomp}). GR00T is therefore sensitive to grip force, and even at the rated force it stays far below its appearance-only cell. For X-VLA and $\pi_{0.5}$ the tested force reduction does not resolve the physical deficit; geometry, length, stroke, inertia and, where read, the gripper-state input remain.

\textbf{Remedies.} Both corrections leave a gap to the training gripper: moving the wrist camera by 6\,cm recovers at most \CoreWxMax{} of 120, and GripSwap, which repaints the new gripper in the wrist image as the training gripper (a Mirage-style cross-painting~\cite{chen2024mirage}), at most \CoreGsMax{} (Table~\ref{tab:core}). Repainting leaves the physical gripper unchanged, including its length and thus the camera--object distance at grasp. The physics-only cell is a reference for removing the appearance mismatch while retaining target physics, not a proven ceiling on image correction (Table~\ref{tab:decomp}). DT-Comp (Sec.~\ref{sec:corr}) comes close to this reference for X-VLA: 39, 50 and 75 of 100 on 2FG7, Hand-E and PGC-50-35, against P = 43, 59 and 81, respectively.

\textbf{Arms.} Replacing the arm alone costs the wrist-camera policies at most \CoreArmLossMax\% of their successes but leaves OpenVLA \CoreOvArm{} of \CoreOvBase. The same holds for the RB5-850e and the Doosan M1013 in the diagnosis sweep (LIBERO-Spatial, 20 episodes): with any of the three arms the wrist-camera policies keep \ArmMiniMin--\ArmMiniMax{} of 20 (nominal \ArmMiniBaseMin--\ArmMiniBaseMax, no significant drop), while OpenVLA falls from \ArmMiniOvBase{} to \ArmMiniOv{} with each arm. 

\begin{table}[t]
\centering
\caption{Successes of 100 across grippers and policies. A/P: appearance-only/physics-only decomposition; None: uncorrected; Both: DT-Comp after bracket. Bold marks the correction selected by lower A (DT-Comp) or P (bracket). Same-run Panda baselines in both runs are X-VLA 91, $\pi_{0.5}$ 85, GR00T 95 and $\tau_0$ 96.}
\label{tab:select}
\scriptsize
\setlength{\tabcolsep}{4pt}
\resizebox{\linewidth}{!}{\begin{tabular}{@{}llccccc@{}}
\toprule
Gripper & Policy & A/P & None & DT-Comp & Bracket & Both \\
\midrule
2FG7 & X-VLA & 0/43 & 0 & \textbf{39} & 14 & 33 \\
 & $\pi_{0.5}$ & 2/35 & 1 & \textbf{39} & 30 & 31 \\
 & GR00T & 72/29 & 17 & 22 & \textbf{65} & 66 \\
 & $\tau_0$ & 49/9 & 17 & 8 & \textbf{42} & 38 \\
\midrule
Hand-E & X-VLA & 0/59 & 0 & \textbf{50} & 3 & 16 \\
 & $\pi_{0.5}$ & 0/48 & 0 & \textbf{31} & 5 & 22 \\
 & GR00T & 68/56 & 44 & 55 & \textbf{75} & 57 \\
 & $\tau_0$ & 27/25 & 9 & 18 & \textbf{45} & 40 \\
\midrule
PGC-50-35 & X-VLA & 0/81 & 0 & \textbf{75} & 0 & 72 \\
 & $\pi_{0.5}$ & 5/52 & 3 & \textbf{46} & 11 & 50 \\
 & GR00T & 55/73 & 46 & \textbf{67} & 46 & 66 \\
 & $\tau_0$ & 63/49 & 9 & 27 & \textbf{42} & 63 \\
\bottomrule
\end{tabular}
}
\end{table}

\subsection{Does the decomposition predict the better correction?}\label{sec:corr}
\textbf{Protocol.}  We test four wrist-camera policies and three grippers on LIBERO-Spatial, with 100 episodes per cell and disjoint initial states for correction and decomposition. Before execution, we fixed the rule that lower A selects DT-Comp and lower P selects the bracket, with ties selecting neither, and one-sided Fisher tests with Holm adjustment per policy--gripper pair; details are in the supplementary sections ``Registration records and chronology'' and ``Confirmatory tests''.

\textbf{Correction ordering.} The correction indicated by the decomposition label scores higher in all 12 original pairs (Table~\ref{tab:select}, Fig.~\ref{fig:grip}(b)). X-VLA and $\pi_{0.5}$ are appearance-led on all three grippers (Fig.~\ref{fig:grip}(a)). GR00T is physics-led on 2FG7 and Hand-E but appearance-led on PGC-50-35, and the better correction changes with its label: the bracket beats DT-Comp by 75 to 55 on Hand-E, whereas DT-Comp beats the bracket by 67 to 46 on PGC-50-35. Reusing the better correction for each policy on 2FG7 identifies the better correction in 7 of 8 added pairs, missing this change. $\tau_0$ is physics-led on all three in the original run. This ordering compares success counts; the paired tests between the two corrections are exploratory and are reported in the supplement. Relative to None, the selected correction raises X-VLA on PGC-50-35 from 0 to 75, $\pi_{0.5}$ on 2FG7 from 1 to 39, and GR00T on 2FG7 from 17 to 65.

\textbf{Image correction methods.} On 2FG7, rendering the training gripper with DT-Comp yielded more successes than the static GripSwap template for $\pi_{0.5}$; the reverse held for $\tau_0$. The differences for X-VLA and GR00T were inconclusive. The two implementations differ in rendering, opening mapping, pasted parts and mask source. The supplement gives counts and paired tests for each policy.

\textbf{Distance to the training gripper.} All selected corrections improve significantly under the registered tests but remain below the same-run Panda baselines (Table~\ref{tab:select}).

\textbf{Combination.} Applying both corrections is not consistently better than the selected single correction. The largest benefit occurs for $\tau_0$ on PGC-50-35: the combination reaches 63, versus 42 with the bracket and 27 with DT-Comp. For X-VLA on Hand-E, however, the combination yields 16 successes against 50 with DT-Comp.

\textbf{Replication.} Two further 100-episode batches test $\tau_0$ and GR00T on Hand-E and PGC-50-35. The numerical ordering of the two corrections holds in all eight batches, and three decomposition labels reproduce. For $\tau_0$ on Hand-E, A/P reverses from 27/25 originally to 31/34 and 30/36. The bracket stays better, so there the decomposition label does not identify the better correction.

\textbf{Appearance-only control.} In the correction run, 2FG7 appearance on Panda physics gives 0, 0, 62 and 49 of 100 without correction for X-VLA, $\pi_{0.5}$, GR00T and $\tau_0$, and 92, 76, 92 and 85 with DT-Comp. On the distinct decomposition initial states, cells A score 0, 2, 72 and 49 (Table~\ref{tab:decomp}). With training-gripper physics, the image correction removes most of the appearance effect. Three of four policies meet the registered baseline criterion, requiring significant improvement and a non-significant two-sided Fisher comparison with Panda; non-significance is not equivalence. Post hoc paired 90\% intervals lie within $\pm10$ percentage points for X-VLA and GR00T only.

\textbf{Budget.} We subsample the eight added-gripper pairs at 40 episodes per policy--gripper pair (two cells of 20). Averaged over the pairs, a direct comparison of the two corrections agrees with the full-data correction ordering in 0.97 of draws, whereas a decomposition of the same size agrees with its full-data label in 0.88. These quantities measure agreement with the observed full-data outcome rather than prospective accuracy.

The twelve original pairs share policies and grippers and are not independent.

\begin{figure*}[t]
\centering
\includegraphics[width=\textwidth]{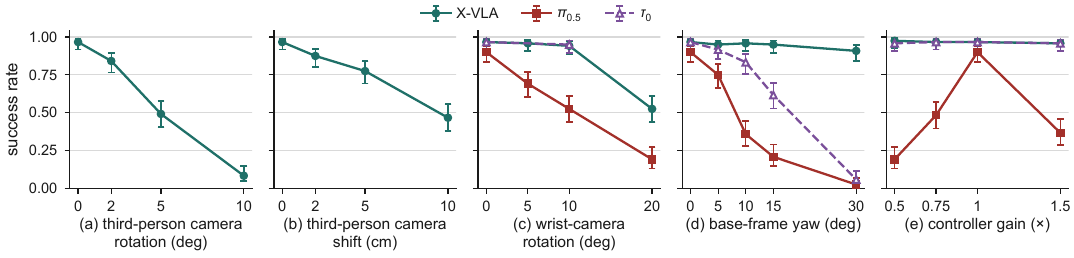}
\caption{Tolerance curves show success rates for X-VLA, $\pi_{0.5}$ and $\tau_0$ across 4 suites, with 120 episodes per point and Wilson 95\% intervals, under third-person camera rotation and shift, wrist-camera rotation, base-frame yaw and controller gain. Third-person camera sweeps cover only X-VLA; the wrist-camera rotation sweep for $\tau_0$ extends to $10^\circ$.}
\label{fig:tol}
\end{figure*}

\subsection{Frame and gain errors need different fixes}\label{sec:action}
The probes identify the injected errors: base-frame yaws of $5^\circ$, $10^\circ$, $15^\circ$ and $30^\circ$ are recovered as $4.5^\circ$, $9.5^\circ$, $14.5^\circ$ and $29.7^\circ$ (median over ten start states), and a gain change appears in the singular values of $C$ while yaw-only setups stay at the noise floor. Table~\ref{tab:action} reports four-suite results. X-VLA, which outputs absolute targets, is insensitive to base-frame yaw and gain. $\pi_{0.5}$ with its stored 50-step chunks collapses under both. Short execution (10 steps per chunk) absorbs the tested gain errors but not frame errors, and the same holds for $\tau_0$, which executes 8 of 10 predicted actions: it keeps 115/120 at half gain but falls to 7/120 at $30^\circ$ yaw. Without retuning, \RM{} raises success under frame errors for both policies, and at $15^\circ$ it lifts all four relative-action policies from \CoreYawMin--\CoreYawMax{} to \CoreRmMin--\CoreRmMax{} of 120 (Table~\ref{tab:core}). Its success count is within \RmOrcMaxDiff{} successes of the oracle over 120 episodes at all reported angles for $\pi_{0.5}$ and $\tau_0$ and at $15^\circ$ for GR00T. The oracle inverts the injected yaw exactly. With short execution at $15^\circ$, both yield \PiOrcNtenFifteen--116 successes out of 120 for $\pi_{0.5}$, never below its nominal success count. For OpenVLA, which executes single-step actions, the probes recover the angle ($14.4^\circ$) but the corrected policy reaches \OvRm{} against the oracle's \OvOrc{} (Fisher $p=\OvRmOrcP$).

\begin{table}[t]
\centering
\caption{Action-side errors, four suites with 120 episodes in total (successes). RM: \RM; oracle: the injected yaw is known and inverted exactly (defined for frame errors only, n/a otherwise); $H{=}10$: execute 10 actions per chunk.}
\label{tab:action}
\scriptsize
\setlength{\tabcolsep}{2.6pt}
\resizebox{\linewidth}{!}{\begin{tabular}{@{}l r r r r r r r r r@{}}
\toprule
 & X-VLA & \multicolumn{5}{c}{$\pi_{0.5}$ (chunk 50)} & \multicolumn{3}{c}{$\tau_0$-VLA (8 of 10)} \\
\cmidrule(lr){3-7}\cmidrule(l){8-10}
Setup & none & none & RM & oracle & $H{=}10$ & RM+$H{=}10$ & none & RM & oracle \\
\midrule
Nominal & 116 & 108 & 108 & n/a & 114 & 114 & 116 & 114 & n/a \\
Yaw $5^\circ$ & 114 & 90 & 110 & 108 & 109 & 115 & 110 & 117 & 115 \\
Yaw $10^\circ$ & 115 & 43 & 109 & 108 & 104 & 114 & 100 & 112 & 115 \\
Yaw $15^\circ$ & 114 & 25 & 110 & 108 & 83 & 116 & 74 & 114 & 116 \\
Yaw $30^\circ$ & 109 & 3 & 110 & 107 & 29 & 116 & 7 & 115 & 116 \\
Gain $\times0.5$ & 117 & 23 & 80 & n/a & 109 & 117 & 115 & 114 & n/a \\
Gain $\times0.75$ & 116 & 58 & 113 & n/a & 112 & 116 & 116 & 116 & n/a \\
Gain $\times1.5$ & 115 & 44 & 104 & n/a & 114 & 113 & 115 & 113 & n/a \\
\bottomrule
\end{tabular}
}
\end{table}

\begin{figure}[t]
\centering
\includegraphics[width=\linewidth]{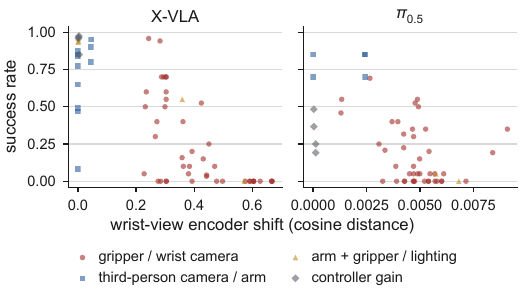}
\caption{Wrist-view encoder shift (cosine distance) versus success for X-VLA and $\pi_{0.5}$; horizontal scales differ. Each point is a setup with 20 episodes (LIBERO-Spatial sweep) or 120 episodes (four-suite tolerance and core setups). Marker classes are gripper / wrist camera, third-person camera / arm, arm + gripper / lighting, and controller gain. The arm + gripper / lighting class changes both views: UR5e with 2FG7 for both policies, and lighting for X-VLA only.}
\label{fig:diag}
\end{figure}

\begin{table}[t]
\centering
\caption{Rollout-free diagnosis. CI: confidence interval; $\rho$: Spearman correlation between the encoder shift and the change in success; LOSO: leave-one-setup-out detection accuracy of a significant drop; majority: accuracy of always predicting the majority class.}
\label{tab:diag}
\scriptsize
\setlength{\tabcolsep}{3pt}
\resizebox{\linewidth}{!}{\begin{tabular}{@{}l r r r r r@{}}
\toprule
Policy & setups & $\rho$ [95\% CI] & LOSO acc. & majority & view attrib. \\
\midrule
X-VLA & 63 & -0.76 [-0.85, -0.62] & 0.87 & 0.79 & 75\% (48) \\
$\pi_{0.5}$ & 54 & -0.55 [-0.72, -0.33] & 0.76 & 0.91 & 88\% (43) \\
OpenVLA$^\ddagger$ & 31 & 0.02 [-0.38, 0.44] & 0.61 & 0.94 & -- \\
GR00T N1.7 & 49 & -0.54 [-0.74, -0.28] & 0.69 & 0.80 & 73\% (37) \\
$\tau_0$-VLA & 49 & -0.61 [-0.76, -0.40] & 0.76 & 0.80 & 76\% (37) \\
\bottomrule
\end{tabular}
}
\\[2pt]
\parbox{\linewidth}{\scriptsize $^\ddagger$Third-person view only (no wrist camera).}
\end{table}

\subsection{Cameras: sensitivity differs between policies}\label{sec:cams}
Camera sensitivity depends on the policy and view (Fig.~\ref{fig:tol}, four suites). For X-VLA, the third-person camera is the sensitive view: rotating it by $2^\circ$ and $10^\circ$ leaves \XCamRotTwo{} and \XCamRotTen{} of 120 successes, and shifting it by 10\,cm leaves \XCamShiftTen, in line with LIBERO-Plus~\cite{fei2025liberoplus}. The wrist camera separates the policies: a $10^\circ$ rotation leaves X-VLA and $\tau_0$ almost unaffected (\XWristRot{} and \TauWristRot) but costs $\pi_{0.5}$ about 40\% of its nominal successes (\PiWristRot; \PiWristRotFive{} already at $5^\circ$ and \PiWristRotTwenty{} at $20^\circ$).

\subsection{Encoder shift ranks visual risk but misses physical failures}\label{sec:diag}
For X-VLA, rotating the external camera caused the larger loss in task success, while features from the wrist camera ranked visual risk more accurately.
Among setups built from official models, the wrist-view shift in the policy's own encoder ranks setups by their change in success for X-VLA, $\pi_{0.5}$, GR00T and $\tau_0$ (Fig.~\ref{fig:diag}, Table~\ref{tab:diag}; $\rho=\XDiagRho$, $\PiDiagRho$, $\GrDiagRho$ and $\TauDiagRho$ over \XDiagN, \PiDiagN, \GrDiagN{} and \TauDiagN{} setups), and it attributes the changed camera correctly in most cases. As a binary detector of a significant drop it exceeds the majority baseline only for X-VLA (0.87 versus 0.79; Table~\ref{tab:diag}), supporting its use for risk ranking and view attribution rather than pass/fail decisions. The ranking comes from the wrist view: on the same episodes, the third-person shift alone gives $\rho=\AgRhoX$ for X-VLA ($\XDiagRho$ from the wrist view) and only $\AgRhoMin$ to $\AgRhoMax$, not significant, for the other three. Consistently, for OpenVLA, which sees only the third-person view, the shift does not rank setups ($\rho=\OvDiagRho$ $\OvDiagCI$ over \OvDiagN{} setups). Success drops in \OvDiagDrop{} of these setups. The diagnosis sweep uses LIBERO-Spatial with 10 tasks $\times$ 2 initial states per setup; Fig.~\ref{fig:diag} also includes four-suite tolerance and core setups. Encoder frames and success outcomes use each policy's own evaluation stack. Failures under the physics-only swap are not distinguished by this signal: for GR00T and $\tau_0$ the 2FG7-physics/Panda-appearance setup has a wrist-view shift comparable to that of the arm swap (GR00T 0.002 vs.\ 0.004; $\tau_0$ 0.029 vs.\ 0.020), yet it fails while the arm swap does not. Action-side errors are invisible at $t{=}0$ and are covered by the probe signals. 

\section{Discussion and Implications for Deployment}
A first frame and a short motion probe answer different deployment questions. The wrist image helps identify changes in visual input; the probe measures how a command moves the end effector. Choosing a gripper correction takes a further step: comparing task success with each candidate. The reversal on replication shows why the initial swap result alone can lead to a poor choice.

These results support four deployment checks.
\begin{enumerate}
\item \textbf{Match the original evaluation environment.} Reproduce the published number with the authors' seeding, simulator version and inference arguments before attributing failure to hardware (Table~\ref{tab:repro}, Sec.~\ref{sec:eval}). The stored checkpoint configuration can differ from the evaluated one.
\item \textbf{Know the action representation.} For relative actions, use six probes and $C=M^{-1}M_0$, with the training response map as reference, to correct frame errors before rollouts (Table~\ref{tab:action}). Short execution absorbs the tested gain errors but not frame errors; X-VLA's absolute targets were insensitive to both (Table~\ref{tab:action}).
\item \textbf{Compare gripper corrections directly.} Try both corrections on each policy--gripper pair (Table~\ref{tab:select}). DT-Comp as evaluated here requires gripper masks and depth from the simulator.
\item \textbf{Rank visual risk before rollouts.} Use encoder shift and view attribution to prioritise setups for rollout evaluation, not as a pass/fail gate (Table~\ref{tab:diag}). The shift misses physical and action-side failures.
\end{enumerate}

\section{Scope and Limitations}\label{sec:limits}
\emph{Transfer to hardware.} The results describe LIBERO tasks with simulated grippers and arms. Validation on physical robots requires repeating the swaps and measuring task success.

\emph{Camera configuration.} Four of the five policies receive wrist images. OpenVLA uses the external view, so the comparison confounds camera configuration with policy identity.

\emph{Action calibration.} Response matching needs the training setup's response map $M_0$ and an invertible $M$; it was tested for base-frame yaw up to $30^\circ$ and controller gains from $\times0.5$ to $\times1.5$.

\emph{Scene variation.} Lighting had little effect in the separate X-VLA check. The joint effect of scene and hardware changes remains open.

\emph{Gripper corrections.} The bracket result measures the combined effect of its adjustments. Deploying DT-Comp on hardware requires camera calibration, a mask of the gripper in the real image and an estimate of the scene depth behind it.

\section{Conclusion}
The gripper correction that performed best depended on both the policy and the gripper. Direct trials distinguished the two corrections even when the appearance-versus-physics diagnosis changed across batches. For base-frame errors, six probe motions calibrated the commands: at $15^\circ$ yaw, three of the four relative-action policies finished within \RmOrcMaxDiff{} successes of exact frame inversion over 120 episodes. Shorter execution helped with gain errors, while frame errors required a coordinate correction. The next test is whether these choices recover task success on physical robots.

\bibliographystyle{IEEEtran}
\bibliography{references}

\end{document}


\maketitle
\thispagestyle{empty}
\pagestyle{empty}

\section{Registration records and chronology}

The protocols for the 2FG7 tests (13 November 2026), the two additional grippers (17 November) and the replication (20 November) specified the conditions, initial states, labelling rule, hypotheses and multiple-testing families before their respective runs.

A changes gripper appearance while retaining Panda physics. P changes gripper physics while retaining Panda appearance. Fewer successes in A select the wrist-image correction (DT-Comp); fewer in P select the bracket; a tie leaves the choice open.

All diagnostic trials preceded the correction trials for X-VLA, $\pi_{0.5}$ and $\tau_0$. For GR00T, all 200 Hand-E trials and 56 of the 200 PGC-50-35 trials preceded correction. The remaining PGC-50-35 trials (44 appearance, 100 physics) followed correction, so the original label for that pair was computed after the correction outcomes were available. The fixed labelling rule used distinct initial states for diagnosis (30--39) and correction (40--49); both independent replication batches gave the original label.

Original runs used seed 0 and 100 episodes per cell: ten LIBERO-Spatial tasks $\times$ ten initial states. GR00T ran on one RTX 4090. Environment worker count and policy batch size changed together from six to four, so their individual effects cannot be estimated. $\tau_0$ ran on RTX 5080 nodes with EGL rendering. Each episode appears once in the released outcomes.

\begin{table*}[t]
\centering
\caption{Confirmatory correction tests. Successes are out of 100 per cell, in the comparison order defined in Sec.~\ref{sec:tests}. Fisher and exact McNemar values are one-sided and Holm-adjusted within the same policy--gripper family. ``vs Panda'' is the two-sided Fisher comparison of the first cell with same-run Panda. Verdicts follow the registered Fisher procedure; n.s.\ denotes a non-significant improvement test.}
\label{tab:tests}
\scriptsize
\begin{tabular}{lllccccl}
\toprule
Policy & Gripper & Test & Successes & Fisher (Holm) & McNemar (Holm) & vs Panda & Registered verdict \\
\midrule
X-VLA & 2FG7 & H1 & 39 vs 0 & $7.1\times10^{-14}$ & $7.3\times10^{-12}$ & $3.6\times10^{-15}$ & improved \\
X-VLA & 2FG7 & H2 & 14 vs 0 & $1.1\times10^{-4}$ & $1.2\times10^{-4}$ & $2.4\times10^{-30}$ & improved \\
X-VLA & 2FG7 & H3a & 33 vs 14 & $2.5\times10^{-3}$ & $3.1\times10^{-5}$ & $4.2\times10^{-18}$ & improved \\
X-VLA & 2FG7 & H3b & 33 vs 39 & 0.85 & 0.92 & $4.2\times10^{-18}$ & n.s. \\
X-VLA & 2FG7 & H4 & 92 vs 0 & $1.9\times10^{-47}$ & $1.0\times10^{-27}$ & 1.0 & criterion met \\
X-VLA & Hand-E & H1 & 50 vs 0 & $8.9\times10^{-19}$ & $3.6\times10^{-15}$ & $1.4\times10^{-10}$ & improved \\
X-VLA & Hand-E & H2 & 3 vs 0 & 0.25 & 0.25 & $9.8\times10^{-42}$ & n.s. \\
X-VLA & Hand-E & H3a & 16 vs 3 & $4.3\times10^{-3}$ & $3.7\times10^{-4}$ & $9.4\times10^{-29}$ & improved \\
X-VLA & Hand-E & H3b & 16 vs 50 & 1.0 & 1.0 & $9.4\times10^{-29}$ & n.s. \\
X-VLA & PGC-50-35 & H1 & 75 vs 0 & $5.7\times10^{-33}$ & $1.1\times10^{-22}$ & $4.2\times10^{-3}$ & improved \\
X-VLA & PGC-50-35 & H2 & 0 vs 0 & 1.0 & 1.0 & $9.4\times10^{-47}$ & n.s. \\
X-VLA & PGC-50-35 & H3a & 72 vs 0 & $4.5\times10^{-31}$ & $6.4\times10^{-22}$ & $8.7\times10^{-4}$ & improved \\
X-VLA & PGC-50-35 & H3b & 72 vs 75 & 1.0 & 1.0 & $8.7\times10^{-4}$ & n.s. \\
$\pi_{0.5}$ & 2FG7 & H1 & 39 vs 1 & $1.8\times10^{-12}$ & $1.5\times10^{-11}$ & $1.8\times10^{-11}$ & improved \\
$\pi_{0.5}$ & 2FG7 & H2 & 30 vs 1 & $4.0\times10^{-9}$ & $5.6\times10^{-9}$ & $1.7\times10^{-15}$ & improved \\
$\pi_{0.5}$ & 2FG7 & H3a & 31 vs 30 & 1.0 & 1.0 & $5.2\times10^{-15}$ & n.s. \\
$\pi_{0.5}$ & 2FG7 & H3b & 31 vs 39 & 1.0 & 1.0 & $5.2\times10^{-15}$ & n.s. \\
$\pi_{0.5}$ & 2FG7 & H4 & 76 vs 0 & $1.4\times10^{-33}$ & $6.6\times10^{-23}$ & 0.15 & criterion met \\
$\pi_{0.5}$ & Hand-E & H1 & 31 vs 0 & $1.2\times10^{-10}$ & $1.9\times10^{-9}$ & $5.2\times10^{-15}$ & improved \\
$\pi_{0.5}$ & Hand-E & H2 & 5 vs 0 & 0.059 & 0.062 & $1.2\times10^{-33}$ & n.s. \\
$\pi_{0.5}$ & Hand-E & H3a & 22 vs 5 & $1.0\times10^{-3}$ & $1.1\times10^{-4}$ & $7.0\times10^{-20}$ & improved \\
$\pi_{0.5}$ & Hand-E & H3b & 22 vs 31 & 0.95 & 0.96 & $7.0\times10^{-20}$ & n.s. \\
$\pi_{0.5}$ & PGC-50-35 & H1 & 46 vs 3 & $3.3\times10^{-13}$ & $4.5\times10^{-13}$ & $8.0\times10^{-9}$ & improved \\
$\pi_{0.5}$ & PGC-50-35 & H2 & 11 vs 3 & 0.049 & 0.057 & $9.5\times10^{-28}$ & improved \\
$\pi_{0.5}$ & PGC-50-35 & H3a & 50 vs 11 & $3.0\times10^{-9}$ & $4.2\times10^{-9}$ & $1.7\times10^{-7}$ & improved \\
$\pi_{0.5}$ & PGC-50-35 & H3b & 50 vs 46 & 0.34 & 0.26 & $1.7\times10^{-7}$ & n.s. \\
GR00T & 2FG7 & H1 & 22 vs 17 & 0.48 & 0.38 & $2.2\times10^{-28}$ & n.s. \\
GR00T & 2FG7 & H2 & 65 vs 17 & $1.1\times10^{-11}$ & $7.3\times10^{-12}$ & $8.9\times10^{-8}$ & improved \\
GR00T & 2FG7 & H3a & 66 vs 65 & 0.5 & 0.5 & $1.9\times10^{-7}$ & n.s. \\
GR00T & 2FG7 & H3b & 66 vs 22 & $9.2\times10^{-10}$ & $7.8\times10^{-10}$ & $1.9\times10^{-7}$ & improved \\
GR00T & 2FG7 & H4 & 92 vs 62 & $7.9\times10^{-7}$ & $3.4\times10^{-7}$ & 0.57 & criterion met \\
GR00T & Hand-E & H1 & 55 vs 44 & 0.24 & 0.11 & $2.2\times10^{-11}$ & n.s. \\
GR00T & Hand-E & H2 & 75 vs 44 & $2.6\times10^{-5}$ & $3.3\times10^{-6}$ & $1.0\times10^{-4}$ & improved \\
GR00T & Hand-E & H3a & 57 vs 75 & 1.0 & 1.0 & $1.3\times10^{-10}$ & n.s. \\
GR00T & Hand-E & H3b & 57 vs 55 & 0.89 & 0.88 & $1.3\times10^{-10}$ & n.s. \\
GR00T & PGC-50-35 & H1 & 67 vs 46 & $8.4\times10^{-3}$ & $2.1\times10^{-3}$ & $4.0\times10^{-7}$ & improved \\
GR00T & PGC-50-35 & H2 & 46 vs 46 & 1.0 & 1.0 & $3.9\times10^{-15}$ & n.s. \\
GR00T & PGC-50-35 & H3a & 66 vs 46 & $1.0\times10^{-2}$ & $5.5\times10^{-3}$ & $1.9\times10^{-7}$ & improved \\
GR00T & PGC-50-35 & H3b & 66 vs 67 & 1.0 & 1.0 & $1.9\times10^{-7}$ & n.s. \\
$\tau_0$ & 2FG7 & H1 & 8 vs 17 & 1.0 & 1.0 & $1.9\times10^{-41}$ & n.s. \\
$\tau_0$ & 2FG7 & H2 & 42 vs 17 & $2.6\times10^{-4}$ & $7.0\times10^{-6}$ & $6.3\times10^{-18}$ & improved \\
$\tau_0$ & 2FG7 & H3a & 38 vs 42 & 1.0 & 1.0 & $6.3\times10^{-20}$ & n.s. \\
$\tau_0$ & 2FG7 & H3b & 38 vs 8 & $1.0\times10^{-6}$ & $4.7\times10^{-9}$ & $6.3\times10^{-20}$ & improved \\
$\tau_0$ & 2FG7 & H4 & 85 vs 49 & $2.1\times10^{-7}$ & $1.1\times10^{-8}$ & 0.014 & improved \\
$\tau_0$ & Hand-E & H1 & 18 vs 9 & 0.097 & 0.049 & $1.9\times10^{-32}$ & n.s. \\
$\tau_0$ & Hand-E & H2 & 45 vs 9 & $1.8\times10^{-8}$ & $5.8\times10^{-11}$ & $1.7\times10^{-16}$ & improved \\
$\tau_0$ & Hand-E & H3a & 40 vs 45 & 0.8 & 0.95 & $6.5\times10^{-19}$ & n.s. \\
$\tau_0$ & Hand-E & H3b & 40 vs 18 & $1.5\times10^{-3}$ & $1.6\times10^{-5}$ & $6.5\times10^{-19}$ & improved \\
$\tau_0$ & PGC-50-35 & H1 & 27 vs 9 & $1.5\times10^{-3}$ & $9.1\times10^{-4}$ & $3.4\times10^{-26}$ & improved \\
$\tau_0$ & PGC-50-35 & H2 & 42 vs 9 & $1.8\times10^{-7}$ & $4.2\times10^{-9}$ & $6.3\times10^{-18}$ & improved \\
$\tau_0$ & PGC-50-35 & H3a & 63 vs 42 & $2.3\times10^{-3}$ & $9.1\times10^{-4}$ & $3.6\times10^{-9}$ & improved \\
$\tau_0$ & PGC-50-35 & H3b & 63 vs 27 & $7.4\times10^{-7}$ & $3.1\times10^{-7}$ & $3.6\times10^{-9}$ & improved \\
\bottomrule
\end{tabular}

\end{table*}

\section{Confirmatory tests}
\label{sec:tests}

Table~\ref{tab:tests} reports the registered comparisons: H1, DT-Comp $>$ none; H2, bracket $>$ none; H3a, both $>$ bracket; H3b, both $>$ DT-Comp; and H4, 2FG7 look-only plus DT-Comp $>$ look-only. ``Both'' applies DT-Comp after the bracket intervention. One-sided Fisher exact tests use Holm adjustment within each policy--gripper family, with $m=5$ for 2FG7 and $m=4$ otherwise, at $\alpha=0.05$.

The registered verdict requires a significant adjusted improvement test. The registered baseline criterion additionally requires a non-significant two-sided Fisher comparison against same-run Panda, reported as ``vs Panda.'' Non-significance in that comparison does not establish equivalence.

Correction cells within a policy--gripper pair share initial states, so their episode outcomes are paired. The registered Fisher test ignores this pairing and is retained as registered. The exact McNemar column provides a paired sensitivity analysis on the shared episodes, using the same one-sided alternatives and Holm families. Episodes within a task are not independent draws of the task distribution; all reported $p$-values are conditional on the ten tasks.

After adjustment, the two tests give different decisions in two cases. For $\pi_{0.5}$ on PGC-50-35, the bracket comparison meets the threshold with Fisher's test ($p=0.049$) and falls just above it with McNemar's ($p=0.057$). For $\tau_0$ on Hand-E, the image correction meets the threshold with McNemar's test ($p=0.049$); Fisher's gives $p=0.097$. The other decisions agree.

The table below gives the unadjusted two-sided exact McNemar $p$-values comparing DT-Comp with the bracket. For the separate comparison with the static GripSwap template on 2FG7, the last column gives DT-Comp and GripSwap successes per 100 episodes, followed by the paired $p$-value.
\begin{center}\scriptsize\setlength{\tabcolsep}{2pt}
\begin{tabular}{@{}lccc l@{}}
\toprule
Policy & 2FG7 & Hand-E & PGC-50-35 & 2FG7 vs. template \\
\midrule
X-VLA & $1.1\times10^{-5}$ & $1.4\times10^{-14}$ & $5.3\times10^{-23}$ & 39 vs. 38 ($1.0$) \\
$\pi_{0.5}$ & 0.16 & $2.2\times10^{-7}$ & $3.1\times10^{-8}$ & 39 vs. 25 ($0.016$) \\
GR00T & $4.2\times10^{-9}$ & $2.9\times10^{-3}$ & $2.5\times10^{-3}$ & 22 vs. 33 ($0.080$) \\
$\tau_0$ & $1.2\times10^{-10}$ & $4.6\times10^{-7}$ & 0.011 & 8 vs. 30 ($1.1\times10^{-5}$) \\
\bottomrule
\end{tabular}\end{center}

\section{Replication}
\label{sec:rep}

Two further batches tested $\tau_0$ and GR00T on Hand-E and PGC-50-35, with 100 episodes per cell. Replication 1 used states 10--19 and seed 1; replication 2 used states 20--29 and seed 2. These states are disjoint from the original states. The reporting rule was fixed before execution, and no further batches were run.

Table~\ref{tab:rep} reports decomposition counts, resulting decomposition labels and correction outcomes. The original DT-Comp--bracket ordering holds in all eight replicated batch--pair combinations. Three policy--gripper decomposition labels reproduce in both batches: $\tau_0$ on PGC-50-35 and GR00T on both grippers. For $\tau_0$ on Hand-E, the decomposition label reverses from physics-led to appearance-led in both batches and then selects the worse correction: DT-Comp scores below the bracket in each batch.

\begin{table*}[t]
\centering
\caption{Original and replication outcomes, with 100 episodes per cell. A/P gives decomposition success counts; appearance-led pairs select DT-Comp and physics-led pairs select the bracket. Paired $p$-values compare DT-Comp and bracket using unadjusted two-sided exact McNemar tests.}
\label{tab:rep}
\scriptsize
\begin{tabular}{lllccccc}
\toprule
Policy & Gripper & Batch & A/P & Label & DT-Comp & Bracket & Paired $p$ \\
\midrule
$\tau_0$ & Hand-E & original & 27/25 & physics & 18 & 45 & $4.6\times10^{-7}$ \\
$\tau_0$ & Hand-E & replication 1 & 31/34 & appearance & 23 & 39 & $3.7\times10^{-3}$ \\
$\tau_0$ & Hand-E & replication 2 & 30/36 & appearance & 25 & 46 & $1.9\times10^{-5}$ \\
$\tau_0$ & PGC-50-35 & original & 63/49 & physics & 27 & 42 & 0.011 \\
$\tau_0$ & PGC-50-35 & replication 1 & 66/50 & physics & 27 & 35 & 0.22 \\
$\tau_0$ & PGC-50-35 & replication 2 & 61/57 & physics & 31 & 41 & 0.099 \\
GR00T & Hand-E & original & 68/56 & physics & 55 & 75 & $2.9\times10^{-3}$ \\
GR00T & Hand-E & replication 1 & 75/59 & physics & 52 & 73 & $1.0\times10^{-4}$ \\
GR00T & Hand-E & replication 2 & 69/50 & physics & 57 & 68 & 0.11 \\
GR00T & PGC-50-35 & original & 55/73 & appearance & 67 & 46 & $2.5\times10^{-3}$ \\
GR00T & PGC-50-35 & replication 1 & 61/74 & appearance & 62 & 46 & 0.02 \\
GR00T & PGC-50-35 & replication 2 & 58/75 & appearance & 73 & 48 & $1.1\times10^{-4}$ \\
\bottomrule
\end{tabular}

\end{table*}

\section{Budget subsampling}
\label{sec:budget}

For each of the eight added-gripper pairs, we draw $k$ initial states per task without replacement, independently across tasks, for $k\in\{1,2,3,5\}$. Each draw uses the same states for the two cells being compared and contains $10k$ episodes per cell. We use 20,000 draws. The column headers 20/40/60/100 in Table~\ref{tab:budget} denote episodes per pair of cells.

``Dec.'' estimates the probability that the subsampled A/P decomposition gives the full-data decomposition label. ``Dir.'' estimates the probability that subsampled DT-Comp and bracket outcomes rank the corrections in the same order as the full data. Ties count as failures for both quantities. ``Boot.'' estimates the probability that a paired bootstrap resample, with replacement, of the full 100-episode decomposition gives the same decomposition label.

The probabilities refer to agreement with the observed full-data results. Diagnosis and correction use separate initial states.

\begin{table*}[t]
\centering
\caption{Agreement with full-data labels and correction ordering under reduced episode budgets. Headers give episodes per pair of cells; Dec., Dir.\ and Boot.\ are defined in Sec.~\ref{sec:budget}. A/P gives full-data decomposition success counts.}
\label{tab:budget}
\scriptsize
\begin{tabular}{lllc cc cc cc cc}
\toprule
 & & & & \multicolumn{2}{c}{20} & \multicolumn{2}{c}{40} & \multicolumn{2}{c}{60} & \multicolumn{2}{c}{100} \\
Policy & Gripper & A/P & Boot. & Dec. & Dir. & Dec. & Dir. & Dec. & Dir. & Dec. & Dir. \\
\midrule
X-VLA & Hand-E & 0/59 & 1.00 & 1.00 & 1.00 & 1.00 & 1.00 & 1.00 & 1.00 & 1.00 & 1.00 \\
X-VLA & PGC-50-35 & 0/81 & 1.00 & 1.00 & 1.00 & 1.00 & 1.00 & 1.00 & 1.00 & 1.00 & 1.00 \\
$\pi_{0.5}$ & Hand-E & 0/48 & 1.00 & 1.00 & 0.95 & 1.00 & 1.00 & 1.00 & 1.00 & 1.00 & 1.00 \\
$\pi_{0.5}$ & PGC-50-35 & 5/52 & 1.00 & 1.00 & 0.97 & 1.00 & 1.00 & 1.00 & 1.00 & 1.00 & 1.00 \\
GR00T & Hand-E & 68/56 & 0.95 & 0.66 & 0.81 & 0.78 & 0.93 & 0.86 & 0.98 & 0.96 & 1.00 \\
GR00T & PGC-50-35 & 55/73 & 1.00 & 0.77 & 0.81 & 0.90 & 0.93 & 0.96 & 0.98 & 1.00 & 1.00 \\
$\tau_0$ & Hand-E & 27/25 & 0.61 & 0.42 & 0.95 & 0.49 & 1.00 & 0.52 & 1.00 & 0.58 & 1.00 \\
$\tau_0$ & PGC-50-35 & 63/49 & 0.97 & 0.70 & 0.73 & 0.83 & 0.87 & 0.91 & 0.95 & 0.99 & 1.00 \\
\midrule
Mean & & & & 0.82 & 0.90 & 0.88 & 0.97 & 0.91 & 0.99 & 0.94 & 1.00 \\
\bottomrule
\end{tabular}

\end{table*}

\section{Paired comparison with Panda}
\label{sec:panda}

Table~\ref{tab:equiv} provides a post hoc paired comparison with same-run Panda over the 100 shared episodes. Let $n_{10}$ count episodes where the evaluated cell succeeds and Panda fails, and $n_{01}$ count episodes where the cell fails and Panda succeeds. With $n=100$, the paired difference in success proportion and its approximate 90\% Wald interval are
\begin{equation}
d=\frac{n_{10}-n_{01}}{n},
\quad
d\ \pm\ \frac{1.645}{n}
\sqrt{n_{10}+n_{01}-\frac{(n_{10}-n_{01})^2}{n}}.
\end{equation}
The interval assumes independent episode pairs and is approximate near the boundary. Differences and interval endpoints in the table are expressed in percentage points.

A cell is read as ``inside $\pm10$'' when the whole interval lies within $\pm10$ percentage points, corresponding to two one-sided tests at 5\%. The post hoc margin corresponds to one success in ten. Readings are pointwise, with no multiplicity adjustment. ``Below Panda'' denotes an interval wholly below zero. These readings are distinct from the registered baseline criterion in Sec.~\ref{sec:tests}.

\begin{table*}[t]
\centering
\caption{Post hoc paired comparisons with same-run Panda. Success counts are out of 100; differences and 90\% intervals are in percentage points. ``Selected'' identifies the correction selected by the original decomposition label for each added-gripper pair. The $\pm10$ margin is post hoc, and readings have no multiplicity adjustment.}
\label{tab:equiv}
\scriptsize
\begin{tabular}{llccccl}
\toprule
Policy & Cell & Successes & Panda & Difference & 90\% interval & Reading \\
\midrule
X-VLA & 2FG7 look-only + DT-Comp & 92 & 91 & $+1$ & $[-3.9,+5.9]$ & inside $\pm10$ \\
X-VLA & 2FG7 DT-Comp & 39 & 91 & $-52$ & $[-60.9,-43.1]$ & below Panda \\
X-VLA & 2FG7 bracket & 14 & 91 & $-77$ & $[-84.3,-69.7]$ & below Panda \\
X-VLA & Hand-E DT-Comp (selected) & 50 & 91 & $-41$ & $[-50.3,-31.7]$ & below Panda \\
X-VLA & PGC-50-35 DT-Comp (selected) & 75 & 91 & $-16$ & $[-22.9,-9.1]$ & below Panda \\
$\pi_{0.5}$ & 2FG7 look-only + DT-Comp & 76 & 85 & $-9$ & $[-16.0,-2.0]$ & below Panda \\
$\pi_{0.5}$ & 2FG7 DT-Comp & 39 & 85 & $-46$ & $[-55.7,-36.3]$ & below Panda \\
$\pi_{0.5}$ & 2FG7 bracket & 30 & 85 & $-55$ & $[-64.1,-45.9]$ & below Panda \\
$\pi_{0.5}$ & Hand-E DT-Comp (selected) & 31 & 85 & $-54$ & $[-62.2,-45.8]$ & below Panda \\
$\pi_{0.5}$ & PGC-50-35 DT-Comp (selected) & 46 & 85 & $-39$ & $[-47.7,-30.3]$ & below Panda \\
GR00T & 2FG7 look-only + DT-Comp & 92 & 95 & $-3$ & $[-7.3,+1.3]$ & inside $\pm10$ \\
GR00T & 2FG7 DT-Comp & 22 & 95 & $-73$ & $[-80.7,-65.3]$ & below Panda \\
GR00T & 2FG7 bracket & 65 & 95 & $-30$ & $[-38.2,-21.8]$ & below Panda \\
GR00T & Hand-E bracket (selected) & 75 & 95 & $-20$ & $[-27.4,-12.6]$ & below Panda \\
GR00T & PGC-50-35 DT-Comp (selected) & 67 & 95 & $-28$ & $[-35.7,-20.3]$ & below Panda \\
$\tau_0$ & 2FG7 look-only + DT-Comp & 85 & 96 & $-11$ & $[-16.1,-5.9]$ & below Panda \\
$\tau_0$ & 2FG7 DT-Comp & 8 & 96 & $-88$ & $[-93.3,-82.7]$ & below Panda \\
$\tau_0$ & 2FG7 bracket & 42 & 96 & $-54$ & $[-62.5,-45.5]$ & below Panda \\
$\tau_0$ & Hand-E bracket (selected) & 45 & 96 & $-51$ & $[-59.5,-42.5]$ & below Panda \\
$\tau_0$ & PGC-50-35 bracket (selected) & 42 & 96 & $-54$ & $[-62.5,-45.5]$ & below Panda \\
\bottomrule
\end{tabular}

\end{table*}

\section{Gripper force settings}
Because open/close commands do not specify force, we use each datasheet's rated force (per finger: 2FG7 140\,N, RG2 40\,N, EGK40 75\,N, Hand-E 185\,N, PGC-50-35 50\,N), reached by setting the finger position gain to force/stroke. For research grippers without datasheet forces, we derive WidowX 14\,N and ALOHA 93\,N per finger from ROBOTIS servo stall torque, crank radius and official drive settings (Interbotix pulse-width modulation (PWM) 250 of 885; ALOHA current-based position mode without a current limit), assuming torque proportional to PWM duty cycle and mid-stroke force $\tau/(2r)$, where $\tau$ is torque and $r$ is crank radius.

\section{Image-correction settings}

DT-Comp renders the Panda finger links in a second MuJoCo instance with all other geometry hidden; this twin is a state-synchronised rendering replica, not a dynamics model. At each control step its arm joints match the controlled robot, and its finger joint is set so that the Panda pad-centre distance matches the target gripper's measured pad distance, clipped to the Panda's range. The evaluated setting is wrist-only, flange-aligned and fingers-only; rendering uses the scene lighting without shadows from the twin. The target mask is obtained from the simulator depth difference with and without target-gripper visuals and is filled by TELEA inpainting with a radius of 5 pixels after $3\times3$ dilation. Twin pixels are pasted where the twin's finger mask is set and its depth is less than the gripper-free scene depth.

The static-template baseline (GripSwap) uses TELEA inpainting with radius 5. Offline renders provide target masks and Panda templates at five opening levels; the measured opening selects the nearest level after mapping each gripper's range. It differs from DT-Comp in rendering, opening mapping, pasted parts and mask source, so their comparison evaluates complete pipelines.

\section{Reproducibility}

The release includes the tables and their numerical inputs, analysis scripts, condition definitions and episode outcomes. Task and initial-state identifiers link the paired observations.